\documentclass[sigconf]{acmart}

\usepackage{booktabs} % For formal tables

\usepackage{pifont}% http://ctan.org/pkg/pifont - for checkmarks
\newcommand{\cmark}{\ding{51}}%

\usepackage{tikz}
\usepackage{amssymb}
\usepackage{graphicx}
\usepackage{subcaption}
\usepackage{mathtools}
\usepackage{makecell}
\usepackage[linesnumbered,lined,boxed,commentsnumbered]{algorithm2e}
\usepackage{amssymb}
\usepackage{bbm}

\usepackage[american]{babel}

\usepackage{amsmath,amsfonts,bm}

\def\eqref#1{equation~\ref{#1}}
\def\1{\bm{1}}

\DeclareMathAlphabet{\mathsfit}{\encodingdefault}{\sfdefault}{m}{sl}
\SetMathAlphabet{\mathsfit}{bold}{\encodingdefault}{\sfdefault}{bx}{n}

\newcommand{\ours}{\textsc{DDGK}}

\definecolor{ramiblue}{HTML}{007bd5}
\definecolor{ramigray}{HTML}{a9a9a9}
\definecolor{ramired}{HTML}{c11b24}

\DeclarePairedDelimiterX{\infdivx}[2]{(}{)}{#1\;\delimsize\|\;#2}

\setcopyright{rightsretained}
\copyrightyear{2019}
\acmYear{2019} 
\setcopyright{iw3c2w3}
\acmConference[WWW '19]{Proceedings of the 2019 World Wide Web Conference}{May 13--17, 2019}{San Francisco, CA, USA}
\acmBooktitle{ Proceedings of the 2019 World Wide Web Conference (WWW'19), May 13--17, 2019, San Francisco, CA, USA}
\acmPrice{}
\acmDOI{10.1145/3308558.3313668}
\acmISBN{978-1-4503-6674-8/19/05}

\begin{document}
\title{DDGK: Learning Graph Representations for Deep Divergence Graph Kernels}
%\titlenote{Produces the permission block, and
%  copyright information}
%\subtitle{Extended Abstract}
%\subtitlenote{The full version of the author's guide is available as \texttt{acmart.pdf} document}

\author{Rami Al-Rfou}
%\authornote{}
\affiliation{%
  \institution{Google AI}
  \streetaddress{1600 Amphitheatre Parkway}
  \city{Mountain View}
  \state{CA}
  \postcode{94043}
}
\email{rmyeid@google.com}

\author{Dustin Zelle}
%\authornote{}
\affiliation{%
  \institution{Google AI}
  \streetaddress{111 8th Ave}
  \city{New York}
  \state{NY}
  \postcode{10011}
}
\email{dzelle@google.com}

\author{Bryan Perozzi}
%\authornote{}
\affiliation{%
  \institution{Google AI}
  \streetaddress{111 8th Ave}
  \city{New York}
  \state{NY}
  \postcode{10011}
  }
\email{bperozzi@acm.org}

\renewcommand{\shortauthors}{R. Al-Rfou et al.}

\begin{abstract}

Can neural networks learn to compare graphs without feature engineering?
In this paper, we show that it is possible to learn representations for graph similarity with neither domain knowledge nor supervision (i.e.\ feature engineering or labeled graphs).
We propose Deep Divergence Graph Kernels, an unsupervised method for learning representations over graphs that encodes a relaxed notion of graph isomorphism. 
Our method consists of three parts.  
First, we learn an encoder for each anchor graph to capture its structure.
Second, for each pair of graphs, we train a cross-graph attention network which uses the node representations of an anchor graph to reconstruct another graph.
This approach, which we call \emph{isomorphism attention}, captures how well the representations of one graph can encode another.
We use the attention-augmented encoder's predictions to define a divergence score for each pair of graphs.
Finally, we construct an embedding space for all graphs using these pair-wise divergence scores. 

Unlike previous work, much of which relies on 1) supervision, 2) domain specific knowledge (e.g. a reliance on Weisfeiler-Lehman kernels), and 3) known node alignment, our unsupervised method jointly learns node representations, graph representations, and an attention-based alignment between graphs.

Our experimental results show that Deep Divergence Graph Kernels can learn an unsupervised alignment between graphs, and that the learned representations achieve competitive results when used as features on a number of challenging graph classification tasks.
Furthermore, we illustrate how the learned attention allows insight into the the alignment of sub-structures across graphs.
\end{abstract}



\maketitle

\section{Introduction}

Deep learning methods have achieved tremendous success in domains where the structure of the data is known a priori.
For example domains like speech and language have intrinsic sequential structure to exploit, while computer vision applications have spatial structure (images) and perhaps temporal structure (videos).
In all these cases, our intuition guides us to build models and learning algorithms based on the structure of the data.
For example, translation invariant convolution networks might search for shapes regardless of their physical position in an image, or recurrent neural networks might share a common latent representation of a concept across distant time steps or diverse domains such as languages.
In contrast, graph learning represents a more general class of problems because the structure of the data is free from any constraints.
A neural network model must learn to solve both the desired task at hand (e.g.\ node classification) and to represent the structure of the problem itself -- that of the graph's nodes, edges, attributes, and communities.

\begin{figure}[t]
\includegraphics[width=\columnwidth]{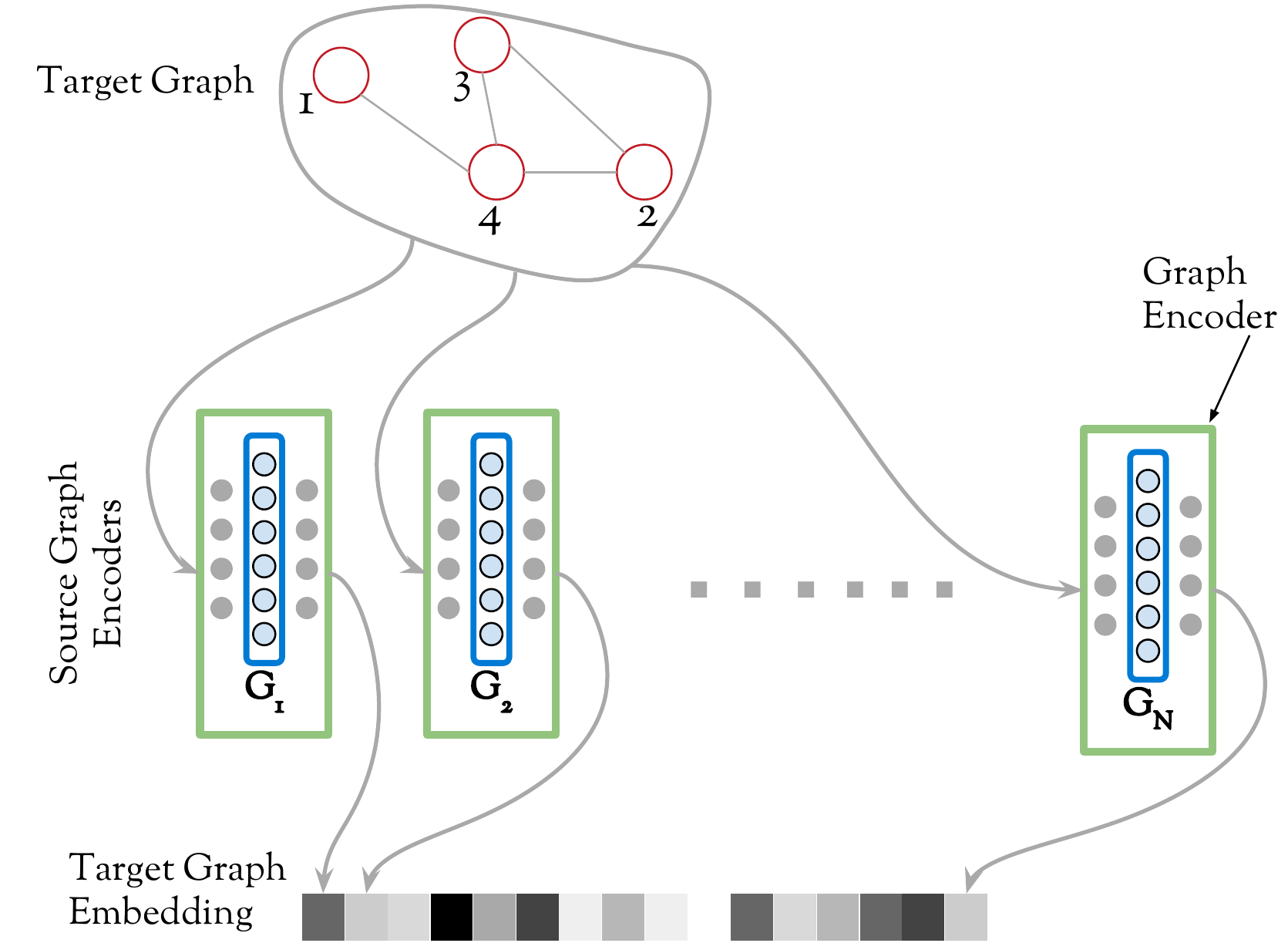}
\caption{Our method of learning graph representations by measuring the divergence of a target graph across a population of source graph encoders.
First, we train a graph encoder for each graph in our source graph population \{$G_1$, $G_2$, ..., $G_N$\}.
Second, for each of these encoders we measure the divergence of the target graph from the associated source graph.
Finally, these divergence scores are used to compose the vector representation of the target graph.}
\label{fig:method}
\end{figure}

Despite the challenges, there has been a recent surge of interest in applying neural network models to such graph-structured data \cite{deepwalk, kipf-gcn,hamilton2017inductive,velickovic2017graph,zhu2018deep}.
While initial approaches like DeepWalk \cite{deepwalk} focused on generic representations of graph primitives (e.g.\ a graph's nodes \cite{deepwalk} or edges \cite{asymmetric}), present approaches ignore learning general graph and node representations in favor of maximizing accuracy on a set of narrow classification tasks.
These approaches, broadly referred to as Graph Neural Networks (GNNs), seek to leverage the structure between data items as a scaffolding to perform computation (e.g. message passing, gradient updates, etc).
The parameters and the activations, use the structure during training, but are tuned primarily to classify the graph's nodes, edges, and/or attributes.

While much effort has focused on unsupervised learning of node representations \cite{deepwalk,node2vec,dngr,tsitsulin2017verse}, edge representations \cite{asymmetric}, or latent community structure \cite{cavallari2017learning,wang2017community,zheng2016node}, relatively little work has focused on the unsupervised learning of representations for entire graphs -- a problem of practical interest in domains such as information retrieval, biology, and natural language processing \cite{gilmer2017neural,battaglia2018relational}.
In cases where GNNs have been applied to the task of learning similarity between graphs, the approaches considered generally come in two flavors: an end-to-end \textit{supervised graph classification} or \textit{graph representation learning}.

In supervised graph classification, the task is to solve an end-to-end whole-graph classification problem (i.e.\ the problem of assigning a label to the entire graph).
These supervised approaches \cite{patchysan,zhang2018end,tixier2018graph,morris2018weisfeiler} learn an intermediate representation of an entire graph as a precondition in order to solve the classification task.
This learned representation can be used to compare similarity between graphs, but is heavily biased towards maximizing performance on the classification task of interest.  

The second class of approaches focuses on the more general problem of learning graph representations \cite{taheri2018RNN}.
While much exciting progress has been made in this area, the existing approaches suffer from one or more of the following limitations.
First, many existing methods rely on feature engineering, such as the graph's clustering coefficient, its motif distribution, or its spectral decomposition, to represent graphs \cite{berlingerio2012netsimile,yanardag2015deep,tsitsulin2018netlsd}.
By limiting the features that they consider, these methods are limited to composing only known graph signals.
Second, many of these approaches \cite{patchysan,zhang2018end} have sought to encode algorithmic heuristics from the graph isomorphism literature (especially the intuition encoded in the Weisfeiler Lehman algorithm \cite{shervashidze2011weisfeiler}).
Relying heavily on existing heuristics to solve a hard problem raises an important question: how well can a learning-only approach solve a classic algorithmic problem?
Finally, other work in this area of graph similarity assumes that identical nodes in both graphs share the same id (i.e.\ the alignment is already given).
While this can be useful for calculating a similarity score, we find the general problem more compelling.

%\todo{find out where to put 1.) graph2vec, 2.) Deep graph kernels, and 3... anyone else in this taxonomy.}

% in this work ....
In this work, we propose a method of learning graph representations driven by the similarity between a pair of graphs as measured by the divergence in their structures.
We show the representations learned through our method, Deep Divergence Graph Kernels (\ours{}), capture the attributes of graphs by using them as features for several classification problems.
In addition, we show that our representations capture the local similarity of graph pairs and the global similarity across families of graphs.

% unlike other methods we ....
\ours{} has three key differentiators.
First, it makes no assumptions about the structure of the matching problem. 
In order to solve the matching problem, we propose an attention mechanism: \emph{isomorphism attention} to align the nodes across graph pairs.
Second, \ours{} does not rely on any existing heuristics for graph similarity.
Instead, we learn the kernel method jointly with the node representation and alignment networks.
This allows the model the freedom to learn representations that best preserve the graph, and does not impose artificial oversights.
Finally, as an unsupervised method, the representations it learns emphasize structural similarity, and does not correlate with a downstream labeling tasks.
This is especially useful for ranking tasks where labeling may not be available.

% our main contributions are ....
\noindent To summarize, our main contributions are:
\begin{itemize}
\item \textbf{Deep Divergence Graph Kernels}: A novel method for learning unsupervised representations of graphs and their nodes.
Our kernel is learnable and does not depend on feature engineering or domain knowledge. 
\item \textbf{Isomorphism Attention}:  A cross-graph attention mechanism to probabilisticly align representations of nodes between graph pairs.
These attention networks allow for great interpretablity of graph structure and discoverablilty of similar substructures.
\item \textbf{Experimental results}: We show that \ours{} both encodes graph structure to distinguish families of graphs, and when used as features, the learned representations achieve competitive results on challenging graph classification problems like predicting the functional roles of proteins.
\end{itemize}

\section{Learning Graph Representations}
In this section, we lay out the problem definition of representing graphs and the connection between our representations and the kernel framework.

\subsection{Problem Definition}

A graph is defined to be a tuple $G=(V, E)$, where $V$ is the set of vertices and $E$ is the set of edges, $E \subseteq (V\times V)$.
A graph $G$ can have an attribute vector $Y$ for each of its nodes or edges.
We denote the attributes of node $v_i$ as $y_i$, and denote the attributes of an edge ($v_i$, $v_j$) as $y_{ij}$.

Given a family of graphs {$G_0$, $G_1$, $\dots$, $G_N$} we aim to learn a continuous representation for each graph $\Psi(G) \in \mathbb{R}^{N}$ that encodes its attributes and its structure.
For this representation to be useful, it has to be comparable to other graph representations.
However, it is likely that our method of graph encoding will produce one of many equally good representations each time we run it.
For example we can get two different, but equal, representations by permuting the dimensions of the first one.
Those representations are not comparable given they exist in two different spaces.

To avoid this problem, we seek to develop an equivalence class across all possible encodings of a graph.
Essentially, two encodings of a graph are equivalent if they lead to the same pair-wise similarity scores when used to compare the graph to all other graphs in the set.
We note that this issue arises when working with embedding based representations across domains, and several equivalence methods have been proposed \cite{D16-1250,NIPS2018_7368}.

\subsection{Embedding Based Kernels}
In this work, we study the development of graph kernels, which are functions to compute the pairwise similarity between graphs.
Specifically, given two graphs $G_1$, $G_2$, a classic example of a kernel defined over graph pairs is the geometric random walk kernel \cite{borgwardt2005protein} as shown in Eq.\ \ref{eq:rwk}:
\begin{equation}
    k_{\times}(G_1, G_2) = e^T(I - \lambda A_\times)^{-1}e,
    \label{eq:rwk}
\end{equation}
where $A_\times$ is the adjacency matrix of the product graph of $G_1$ and $G_2$, and $\lambda$ is a hyper-parameter which encodes the importance of each step in the random walk.
We aim to learn an embedding based kernel function $k()$ as a similarity metric for graph pairs, defined as the following:
\begin{equation}
    k(G_1, G_2) = || \Psi(G_1) - \Psi(G_2) ||^2
    \label{eq:our_kernel}
\end{equation}
For a dataset of $N$ \emph{source}\footnote{In this paper, we use source and anchor interchangeably when referring to the encoded graph.} graphs $\mathcal{S}$ and $M$ \emph{target} graphs ($\mathcal{T})$, for any member of the target graph set we define the $i^{th}$ dimension of the representation $\Psi(G \in \mathcal{T}) \in \mathbb{R}^N$ to be:
\begin{equation}
    \Psi(G)_i = \sum_{v_j \in V_T} f_{g_i}(v_j),
    \label{eq:embedding_dim}
\end{equation}
where $g_i \in \mathcal{S}$ and $f_{g_i}()$ is a predictor of some structural property of the graph $G$ but parameterized by the graph $g_i$.
We note that the source and target graphs sets ($\mathcal{S}, \mathcal{T}$) could be disjoint, overlapping, or equal.

\begin{comment}
\subsection{Representation Properties}
We seek to learn graph representations through a process which has the following desirable
characteristics:

\begin{itemize}
    \item Unsupervised - the distance between graphs should be computable without side information
    \item Unaware (General)
    \item Self-aligning
    \item Can account for attribute information (both node and edge) attributes.
\end{itemize}

We detail the method that we design for this below.
\end{comment}

\section{Learning to Align Graph Representations}
\label{sec:aligngraphs}
% \subsection{Algorithm: Deep Divergence Graph Kernels}
We propose to learn a graph representation by comparing it to a population of graphs.
To compare the similarity of a pair of graphs (\emph{source}, \emph{target}), we rely on deep neural networks to measure the divergence between their structure and attributes.
First, we learn the structure of the source graph by passing it through a graph encoder.
Second, to measure how much the target graph diverges from the source graph,
we use the source graph encoder to predict the structure of the target graph.
If the pair is similar, we expect the source graph encoder to correctly predict the target graph's structure.
%Third, we construct the target graph representation by calculating the divergence scores across a population of source graphs.
%Figure \ref{fig:method} illustrates the construction of the target graph vector representation using a population of source graphs.
In this section, we develop the three key components necessary to learn the similarity between a pair of graphs.

First, in Section \ref{sec:encoding}, we discuss encoding graphs. 
The quality of the graph representation depends on the extent to which the encoder of each source graph is able to discover its structure.

Second, in Section \ref{sec:attention}, we propose a cross-graph attention mechanism to learn a soft alignment between graphs.
This is necessary because a target graph may not share its vertex ids with any of the source graphs -- indeed, they could even have differing number of nodes!
Therefore, we need to learn an alignment between the nodes of the target graph and each source graph.
This leads to an alignment that is not necessarily a one-to-one correspondence.
%Section \ref{sec:attention} shows how to use cross-graph attention to learn such soft alignment.

Third, in Section \ref{sec:attributes} we introduce additional constraints on the cross-graph attention learning.
%the graph attention may not preserve edge or node attributes.
For example, let us assume that $v_i \in V_{G_1}$ is assigned to $u_j \in V_{G_2}$.
While both $v_i$ and $u_j$ may be structurally similar, they may belong to different node classes as indicated by their attributes.
These attributes may be of significant importance to the nature of the graph.  For instance, swapping one element for another in a graph representing a molecule could drastically change its chemical structure.

We will see how these pairwise alignments can produce divergence scores suitable for Graph Kernels in Section \ref{sec:graph_kernels}.

\begin{figure}[t]
\centering
\includegraphics[width=\columnwidth]{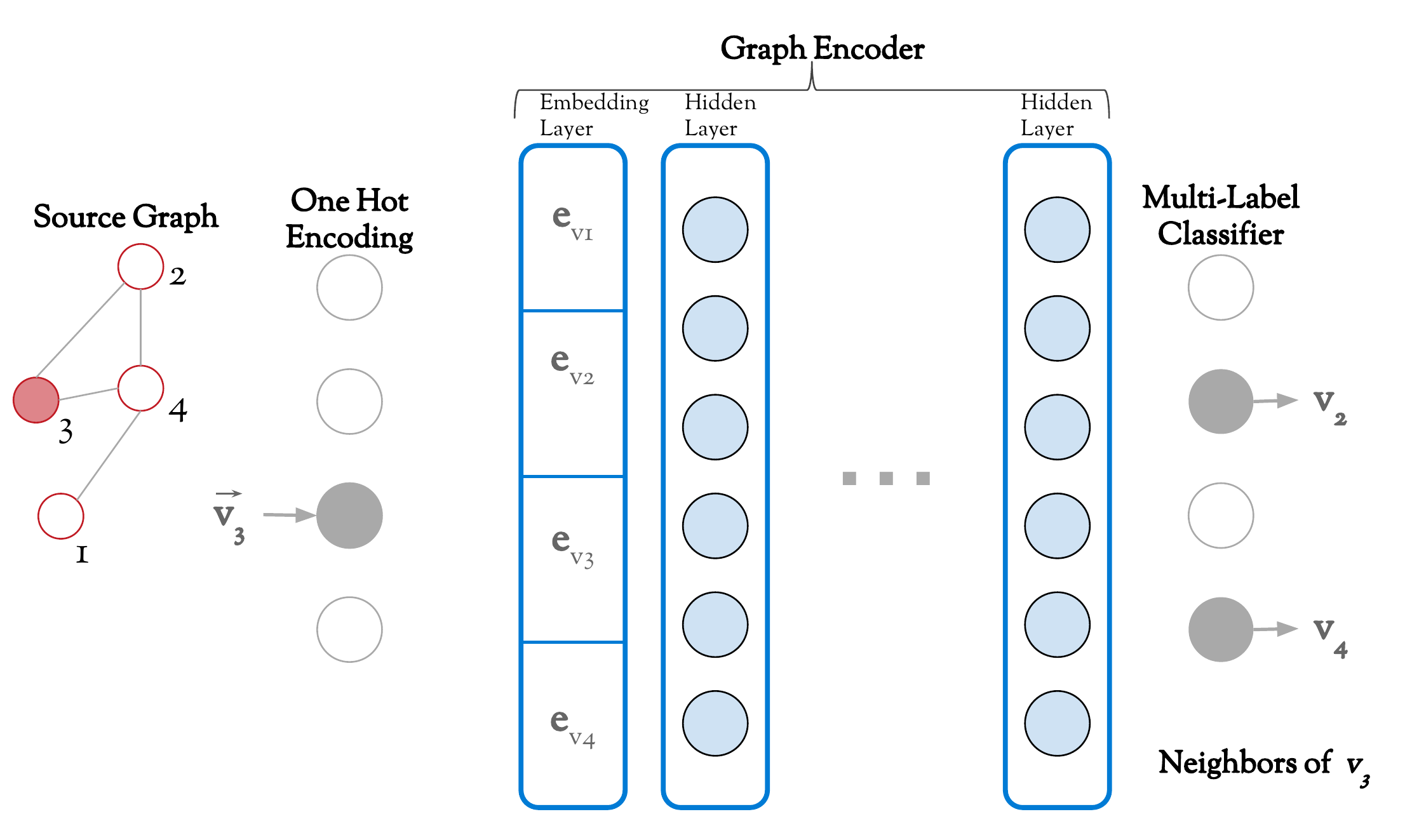}
\caption{A Node-To-Edges Encoder.
Here the input graph contains 4 vertices, and the encoder has to predict the neighbors of vertex $v_3$.
First, $v_3$ is represented by a one-hot encoding $\vec{v}_3$.
Second, $\vec{v}_3$ is multiplied by a linear embedding layer.
Third, this embedding $\mathbf{e}_{v_3}$ is passed to a DNN which produces scores for each vertex in $V$.
Finally, these scores are normalized using the \emph{sigmoid} function to produce the final predictions, in this case, \{$v_2$, $v_4$\}.}
\label{fig:ae}
\end{figure}

\subsection{Graph Encoding}
\label{sec:encoding}
To learn the structure of a graph, we train an encoder capable of reconstructing such structure given partial or distorted information.
In this paper, we choose a \emph{Node-To-Edges} encoder (Figure \ref{fig:ae}) for its simplicity, but we note that additional choices are certainly possible (see Section \ref{sec:extensions} for more discussion).
\paragraph{Node-To-Edges Encoder} -
In this setup, an encoder is given a single vertex and it is expected to predict its neighbors.
This can be modeled as a multilabel classification task since the predictions are not mutually exclusive.
Specifically, we are maximizing the following objective function $J(\theta)$,
\begin{equation}
J(\theta) = \sum_i \sum_{\substack{j \\ e_{ij} \in E}} \log \Pr(v_j \mid v_i, \theta).
\end{equation}
Each vertex $v_i$ in the graph is represented by one-hot encoding vector $\vec{v_i}$.
Then to embed the vertex we multiply its encoding vector with a linear layer $\mathbf{E} \in \mathbb{R}^{|V| \times d}$ resulting in an embedded vertex $\mathbf{e}_{v_i} \in \mathbb{R}^d$, where $|V|$ is the number of vertices in the graph, and $d$ is the size of the embedding space.

For graphs with a large number of nodes, we can replace this multiplication with a table lookup, extracting one row from the embedding matrix.
This embedding vector represents the feature set given to the encoder tasked with predicting all adjacent vertices.
Our encoder $H$, is implemented as a fully connected deep neural network (DNN) with an output layer of size $|V|$ and trained as a multilabel classifier.

\begin{figure}[t]
\includegraphics[width=\columnwidth]{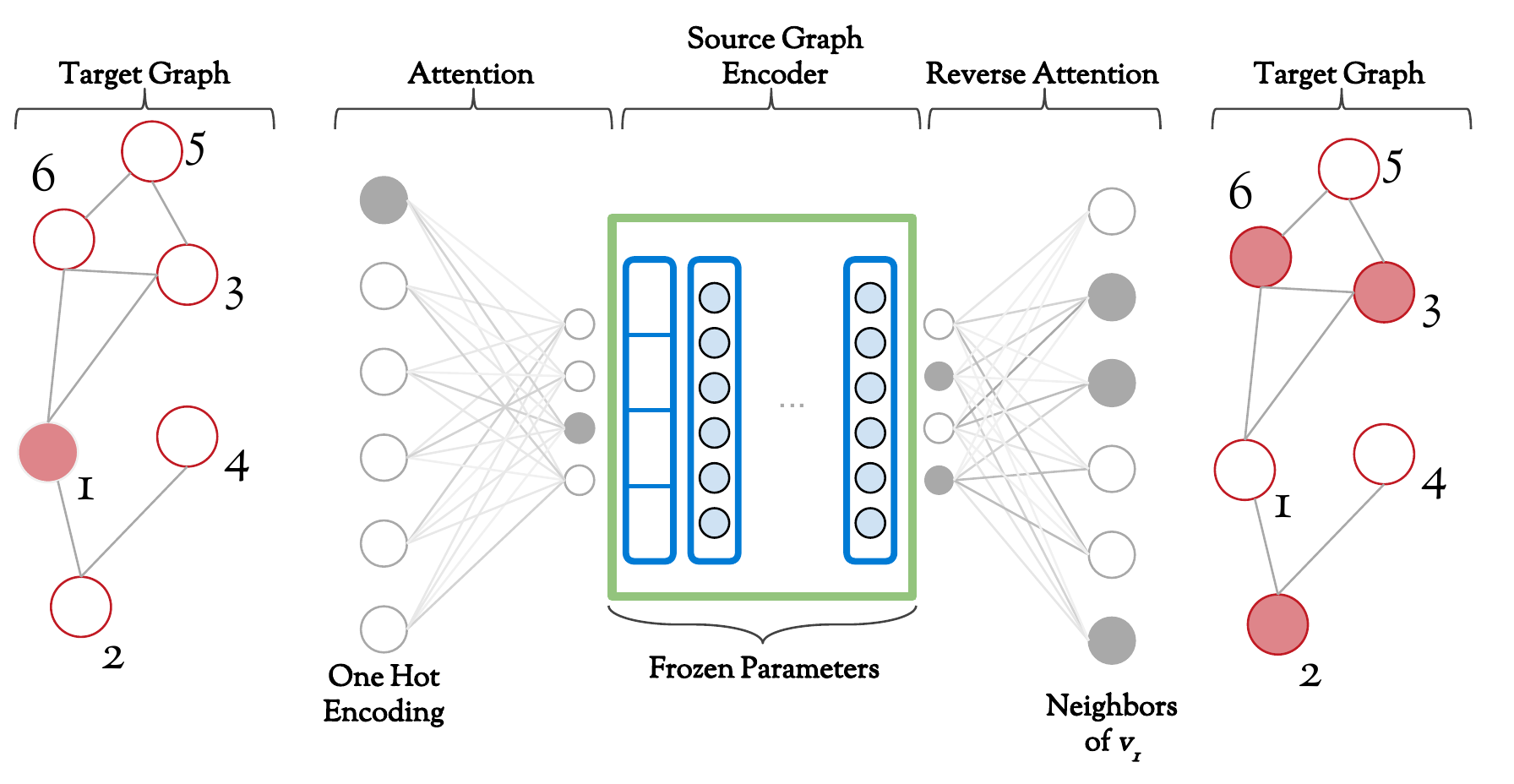}
\caption{Attention layers mapping the target graph nodes onto the source graph.
The augmented encoder has to predict the neighbors of node $1$ in the target graph.
First, node $1$ is passed to the attention layer which assigns it mainly to node $3$ of the source graph.
Second, the source graph encoder learned earlier (in Figure \ref{fig:ae}) that the neighbors of node $3$ are \{$2$, $4$\}.
Finally, the reverse attention network maps nodes \{$2$, $4$\} of the source graph to nodes \{$2$, $3$, $6$\} of the target graph which are the neighbors of node $1$.
}
\label{fig:attention}
\end{figure}

\subsection{Cross-Graph Attention}
\label{sec:attention}
So far, we have developed a utility to encode individual graphs. 
However, we seek to develop a method which can compare pairs of graphs, which may  differ in size (differing node sets) and structure (differing edge sets). 
For this to happen we need a method of learning an alignment between the graphs. Ideally this method will operate in the absence of a direct mapping between nodes.

%Comparing them directly is a difficult problem.

In other areas, attention models have been proposed to align structured data.
For example, attention models have been proposed to align pairs of images and text \cite{xu2015show}, pairs of sentences for translation \cite{vaswani2017attention}, and pairs of speech and transcription \cite{NIPS2015_5847}.
Inspired by these efforts, we formalize the problem of aligning two graphs as that of attention.
We propose an attention mechanism, \emph{isomorphism attention}, that aligns the nodes of a target graph against those of a source graph.
%To tackle this, we introduce an attention mechanism to learn vertex alignments across graphs.

\subsubsection{Isomorphism Attention}
Given two graphs $S$ (\emph{source graph}) and $T$ (\emph{target graph}), we propose a model that allows bi-directional mapping across the pair's nodes.
This requires two separate attention networks.
The first network allows nodes in the target graph to \emph{attend} to the nodes in the source graph.
The second network, allows neighborhood representations in the source graph to \emph{attend} to neighborhoods in the target graph.

We denote the first attention network as ($\mathcal{M}_{T\rightarrow S}$), which assigns every node in the target graph $(u_i \in T)$ a probability distribution over the nodes of the source graph $(v_j \in S)$.
This attention network will allow us to pass the nodes of the target graph as an input to the source graph encoder.
We implement this attention network using a multiclass classifier,
\begin{equation}
\Pr(v_j \mid u_i) = \frac{e^{\mathcal{M}_{T\rightarrow S}(v_j, u_i)}}{\sum_{v_k \in V_S} e^{\mathcal{M}_{T\rightarrow S}(v_k, u_i)}}.
\end{equation}
The second network is a \emph{reverse attention} network ($\mathcal{M}_{S\rightarrow T}$) which aims to learn how to map a neighborhood's representation in the source graph to a neighborhood in the target graph.
By adding both attention networks to the source graph encoder, we will be able to construct a target graph encoder that is able to predict the neighbors of each node -- but utilizing the structure of the source graph.
We implement the reverse attention as a multilabel classifier,
\begin{equation}
\Pr(u_j \mid \mathcal{N}(v_i)) = \frac{1}{1 + e^{- \mathcal{M}_{S\rightarrow T}(u_j, \mathcal{N}(v_i))}}.
\end{equation}
%The network learns a soft-alignment where one vertex in target graph could be mapped to one or more vertices in the source graph.
%Then, this soft alignment get fed to the embedding layer of the source graph.
%The source graph encoder is still predicting nodes in its vertex space of size $|V_P|$.
%Therefore, we need to learn a reverse attention layer that could ``translate" back the prediction into the target graph nodes.
Figure \ref{fig:attention} shows the attention network ($\mathcal{M}_{T\rightarrow S}$) receiving a one-hot encoding vector representing a node ($u_i$) in the target graph and mapping it onto the most structurally similar node ($v_j$) from the source graph.
The source graph encoder, then, predicts the neighbors of $v_j$, $\mathcal{N}(v_j)$.
The \emph{reverse attention} network ($\mathcal{M}_{S\rightarrow T}$), takes $\mathcal{N}(v_j)$ and maps them to the neighbors of $u_i$, $\mathcal{N}(u_i)$.

Both attention networks may be implemented as linear transformations $\mathbf{W}_A \in \mathbb{R}^{|V_Q| \times |V_P|}$.
In the case that either $|V_P|$ or $|V_Q|$ are prohibitively large, the attention network parameters can be decreased by substituting a DNN with hidden layers of fixed size.
This will reduce the attention network size from $\Theta(|V_P| \times |V_Q|)$ to $\Theta(|V_P| + |V_Q|)$.

\subsection{Attributes Consistency}
\label{sec:attributes}
Labeled graphs are not defined only by their structures, but also by the attributes of their nodes and edges.
%For example, you could drastically change the chemical properties of a molecule by swapping one element for another while keeping its structure the same.
The attention network assigns each node in the target graph a probability distribution over the nodes of the source graph.
There might be several, equally good, nodes in the source graph with similar structural features.
However, these nodes may differ in their attributes.
To learn an alignment that preserves nodes and edges attributes, we add regularizing losses to the attention and reverse-attention networks.

More specifically, we refer to the nodes as $v$ and $u$ for the source and target graphs, respectively.
We refer to the set of attributes as $\mathcal{Y}$ and the distribution of attributes over the graph nodes as $(Q_n = \Pr(y_i \mid u))$.
Given that the attention network $\mathcal{M}_{T\rightarrow S}$ learns the distribution $\Pr(u_k \mid v_j)$,
we can calculate a probability distribution over the attributes as inferred by the attention process as the following:
\begin{equation}
Q_n(y_i | u_j) = \sum_k\mathcal{M}_{T\rightarrow S}(y_i | v_k) \Pr(v_k \mid u_j).
\label{eq:attrdistnodes}
\end{equation}

\noindent We define, the attention regularizing loss over the nodes attributes to be the average cross entropy loss between the observed distribution of attributes and the inferred one (See Eq. \ref{eq:lattnodes}).
\begin{equation}
    L = \frac{1}{|V_T|} \sum_j^{|V_T|} \sum_{i} \Pr(y_i \mid u_j) \log(Q_n(y_i | u_j)),
    \label{eq:lattnodes}
\end{equation}
where $|V_T|$ is the number of nodes in the target graph.

For preserving edge attributes over nodes, we define $Q_e(y_i \mid u) = \Pr(y_i \mid u)$ to be the normalized attributes count over all edges connected to the node $u$.
For instance, if a node $u$ has 5 edges with 2 of them colored red and the other three colored yellow, $Q_e(red \mid u) = 0.4$
By replacing $Q_n$ with $Q_e$ in Equations \ref{eq:attrdistnodes} and \ref{eq:lattnodes}, we create a regularization loss for edge attributes.

We also introduce these regularization losses for \emph{reverse attention} networks.
Reverse attention networks maps a neighborhood in the source graph to a neighborhood in the target graph.
The distribution of attributes over a node's neighborhood will be the frequency of each attribute occurrence in the neighborhood normalized by the number of attributes appearing in the neighborhood.
For edges, the node's neighborhood edges are the edges appearing at 2-hops distance from the node.
Similarly, we can define the probability of the edges attributes by normalizing their frequencies over the total number of attributes of edges connected to the neighborhood.

\section{Deep Divergence Graph Kernels}
\label{sec:graph_kernels}

So far, we have proposed a method for learning representations of graphs, and an attention mechanism for aligning graphs based on a set of encoded graph representations. 
Here we discuss our proposed method for using the alignment to construct a graph kernel based on divergence scores.
First, in Section \ref{sec:div}, we show how we can utilize the divergence scores to construct a full graph representation.
Divergence is driven by the target graph structure and attribute prediction error as calculated using a source graph encoder.
Next  we introduce DDGK, our method for learning graph representations for Deep Divergence Graph Kernels in Section \ref{sec:alg_ddgk}.
Then in Section \ref{sec:training} we discuss how we train these representations.
Finally we discuss the scalability of this approach in Section \ref{sec:scalability}.

\subsection{Graph Divergence}
\label{sec:div}
In Section \ref{sec:aligngraphs} we presented a method to align two graphs by using a source graph encoder, augmented with attention layers, to encode a target graph.
Here, we propose to use the ability of the augmented encoder at predicting the structure of the target graph as a measure of those graphs similarity.
To explain, let us assume the trivial case where both the source and target graphs are identical.
First, we train the source graph encoder.
Second, we augment it with attention networks and train it to predict the structure of the target graph.
The attention networks will (ideally) learn the identity function.
Therefore, the source graph encoder is able to encode the target graph as accurately as encoding itself.
We would reasonably conclude that these graphs are similar.

We aim to learn a metric that measures the divergence score between a pair of graphs $\{S, T\}$.
If two graphs are similar, we expect their divergence to be correspondingly low.
We refer to the encoder trained on a graph $S$ as $H_S$ and the divergence score given to the target graph $T$ to be  

\begin{equation}
    \mathcal{D}^\prime\infdivx{T}{S} = \sum_{v_i \in V_T} \sum_{\substack{j \\ e_{ji} \in E_T}} -\log \Pr(v_j \mid v_i, H_S)
\end{equation}

\noindent Given that $H_S$ is not a perfect predictor of the graph $S$ structure, we can safely assume that $\mathcal{D}^\prime\infdivx{S}{S} \neq 0$.
To rectify this problem we define
\begin{equation}
\mathcal{D}\infdivx{S}{T} = \mathcal{D}^\prime\infdivx{S}{T} - \mathcal{D}^\prime\infdivx{S}{S},
\end{equation}
which sets $\mathcal{D}\infdivx{S}{S}$ to zero.

We note that this definition is not symmetric (as $\mathcal{D}\infdivx{T}{S}$ might not necessarily equal to $\mathcal{D}\infdivx{S}{T}$).
If symmetry is required, we can define $\mathcal{D}(S, T) = \mathcal{D}\infdivx{S}{T} + \mathcal{D}\infdivx{T}{S}$.

\subsection{Graph Embedding}
Given a set of source graphs, we can establish a vector space where each dimension corresponds to one graph in the source set.
Target graphs are represented as points in this vector space where the value of the $i_{th}$ dimension for a given target graph $T_j$ is $\mathcal{D}\infdivx{T_j}{S_i}$.

More formally, for a set of $N$ source graphs we can define our target graph representation to be:
\begin{equation}
    \Psi(G_T) = [\mathcal{D}\infdivx{T}{S_0}, \mathcal{D}\infdivx{T}{S_1}, \dots, \mathcal{D}\infdivx{T}{S_N}]
    \label{eq:ddgk_kernell}
\end{equation}

\noindent To create a kernel out of our graph embeddings, we use the Euclidean distance measure as outlined in Eq \ref{eq:our_kernel}.
This distance measure will guarantee a positive definite kernel \cite{haasdonk2004learning,wu2018d2ke}.

\subsection{Algorithm : \textsc{DDGK}}
\label{sec:alg_ddgk}

\IncMargin{1em}
\begin{algorithm}[t!]
\SetKwArray{Encodings}{encodings}
\SetKwArray{Embedding}{$\Psi$}
\SetKwFunction{GraphDivergence}{GraphDivergence}
\SetKwFunction{GraphEncode}{GraphEncode}
\SetKwInOut{Input}{input}
\SetKwInOut{Output}{output}
\Input{
Set of $N$ source graphs $\mathcal{S}$ \\ 
Set of $M$ target graphs $\mathcal{T}$ \\ 
Learning rate $\alpha{}$ \\
Encoding epochs $\tau{}$ \\
Scoring epochs $\rho{}$
}
\Output{All graph representations
$\Psi{} \in \mathbb{R}^{M\times N}$}

\emph{// learn graph encodings}

\ForEach{$g_{i}\in \mathcal{S}$}{
$V, E\leftarrow g_{i}$\\

\For{$step\leftarrow 0$ \KwTo $\tau$}{
$J(\theta) = - \sum_{s} \sum_{\substack{t \\ e_{st}\in E}} \log \Pr(v_t \mid v_s, \theta)$ \label{encoding:loss} \\
$\theta = \theta - \alpha * \frac{\partial J}{\partial \theta}$ \label{encoding:update} \\
}
\Encodings{i}$\leftarrow \theta$\\
}

\ForEach{$g_{i}\in \mathcal{T}$}{
$V, E\leftarrow g_i$ \\
\ForEach{$\theta_j\in \Encodings$}{

\emph{// learn cross-graph attention $\mathcal{M}_{T\rightarrow S}$ and $\mathcal{M}_{T\rightarrow S}$} \\

\For{$step\leftarrow 0$ \KwTo $\rho$}{
$J(\mathcal{M}_{T\rightarrow S}, \mathcal{M}_{S\rightarrow T}) = \scriptstyle{-\sum_s \sum_{\substack{t \\ e_{st}\in E}} \log \Pr(v_t\mid v_s, \theta_j, \mathcal{M}_{T\rightarrow S}, \mathcal{M}_{S\rightarrow T})}$ \label{divergence:loss} \\
$\mathcal{M}_{T\rightarrow S} = \mathcal{M}_{T\rightarrow S} - \alpha * \frac{\partial J}{\partial \mathcal{M}_{T\rightarrow S}}$ \label{divergence:update1} \\
$\mathcal{M}_{S\rightarrow T} = \mathcal{M}_{S\rightarrow T} - \alpha * \frac{\partial J}{\partial \mathcal{M}_{S\rightarrow T}}$ \label{divergence:update2} \\
}

\emph{// calculate graph divergences} \\

\Embedding{i, j}$\leftarrow J(\mathcal{M}_{T\rightarrow S}, \mathcal{M}_{S\rightarrow T})$\\
}
}

\KwRet{\Embedding}
\caption{
DDGK: An unsupervised algorithm for learning graph representations.}
\label{algorithm:ddgk}
\end{algorithm}\DecMargin{1em}

We present pseudo-code for \ours{} in Algorithm \ref{algorithm:ddgk}. 
The algorithm has two parts. First, a \emph{Node-To-Edges} encoder is trained for all source graphs (Algorithm \ref{algorithm:ddgk} line~\ref{encoding:loss} and line~\ref{encoding:update}). 
Second, cross-graph attentions are learned for all target-source graph pairs (Algorithm \ref{algorithm:ddgk} line~\ref{divergence:loss}, line~\ref{divergence:update1} and line~\ref{divergence:update2}). 
We implement \ours{} using a deep neural network for its \emph{Node-To-Edges} encoder and linear transformations for its isomorphism attention. 

\begin{figure*}[t]
    \begin{subfigure}[b]{0.33\textwidth}
        \includegraphics[scale=0.25]{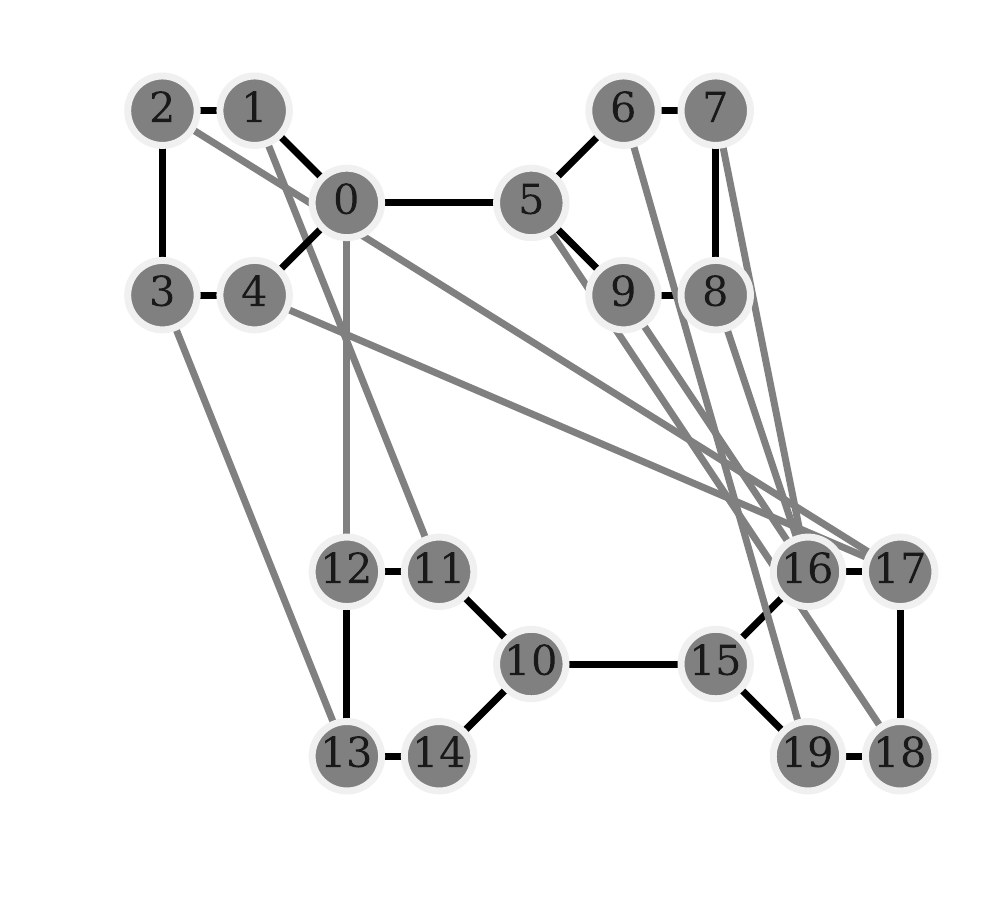} \includegraphics[scale=0.25]{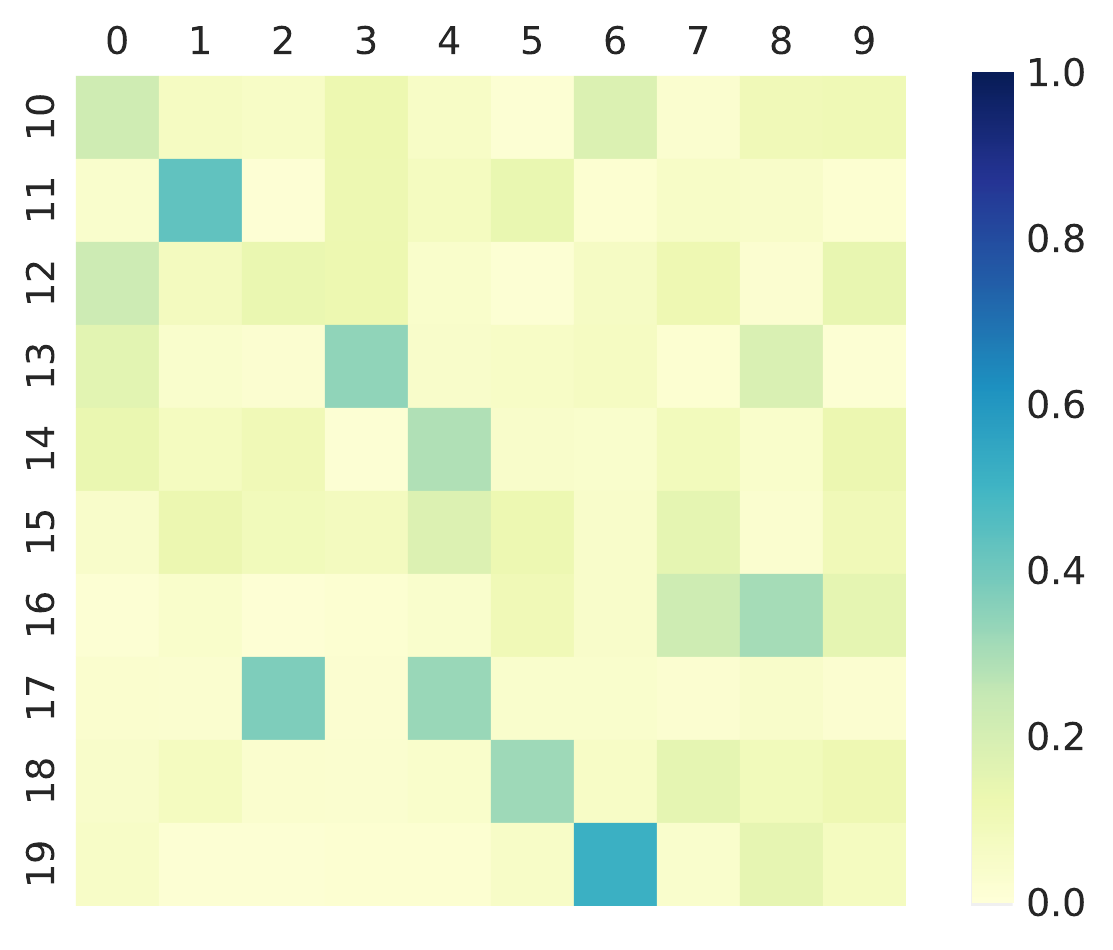}
        \caption{No attributes.}
        \label{fig:graph_no_reg}         
    \end{subfigure}
    \begin{subfigure}[b]{0.33\textwidth}
        \includegraphics[scale=0.25]{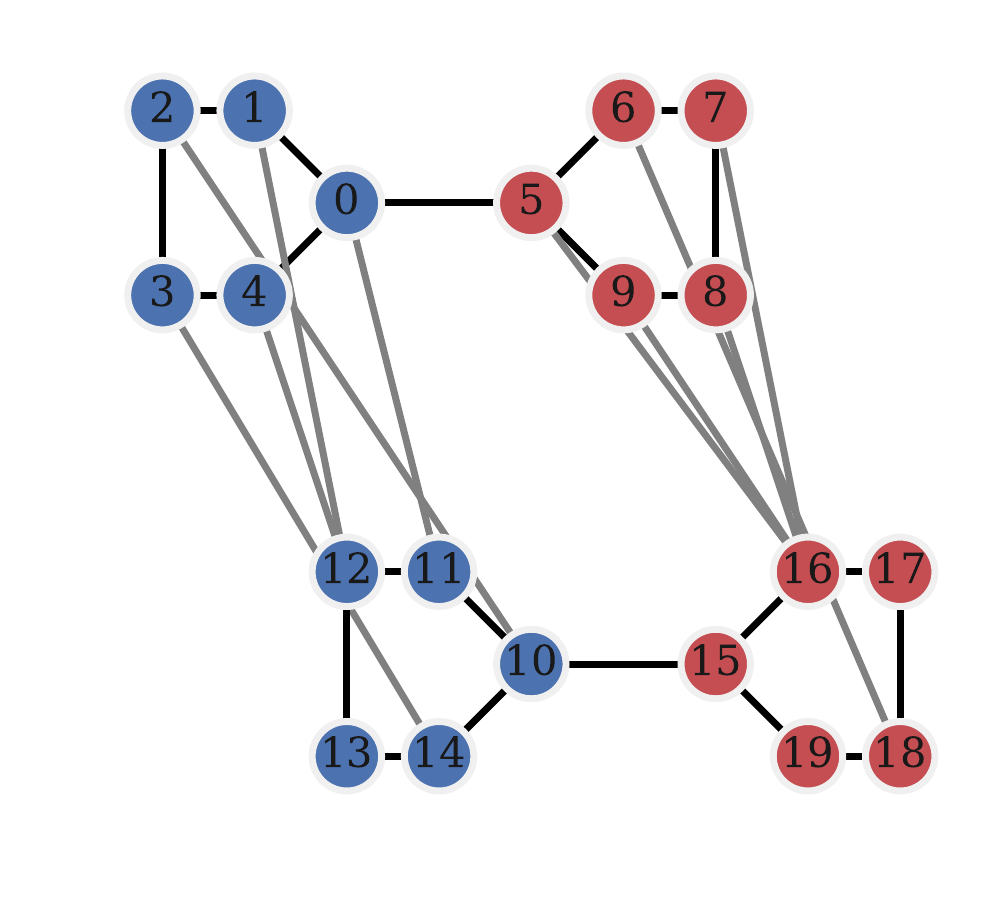}
        \includegraphics[scale=0.25]{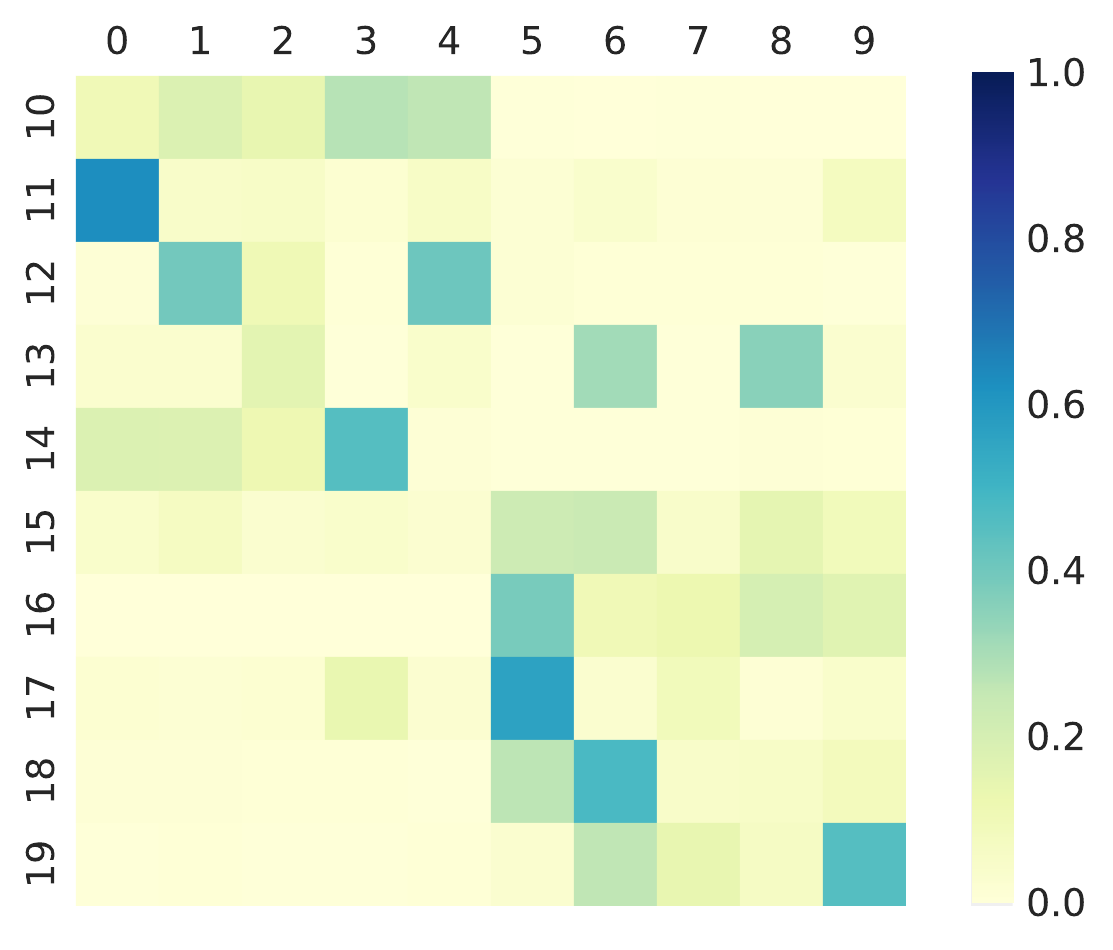}
        \caption{Labeled nodes.}
        \label{fig:graph_node_reg}        
    \end{subfigure}
    \begin{subfigure}[b]{0.33\textwidth}
        \includegraphics[scale=0.25]{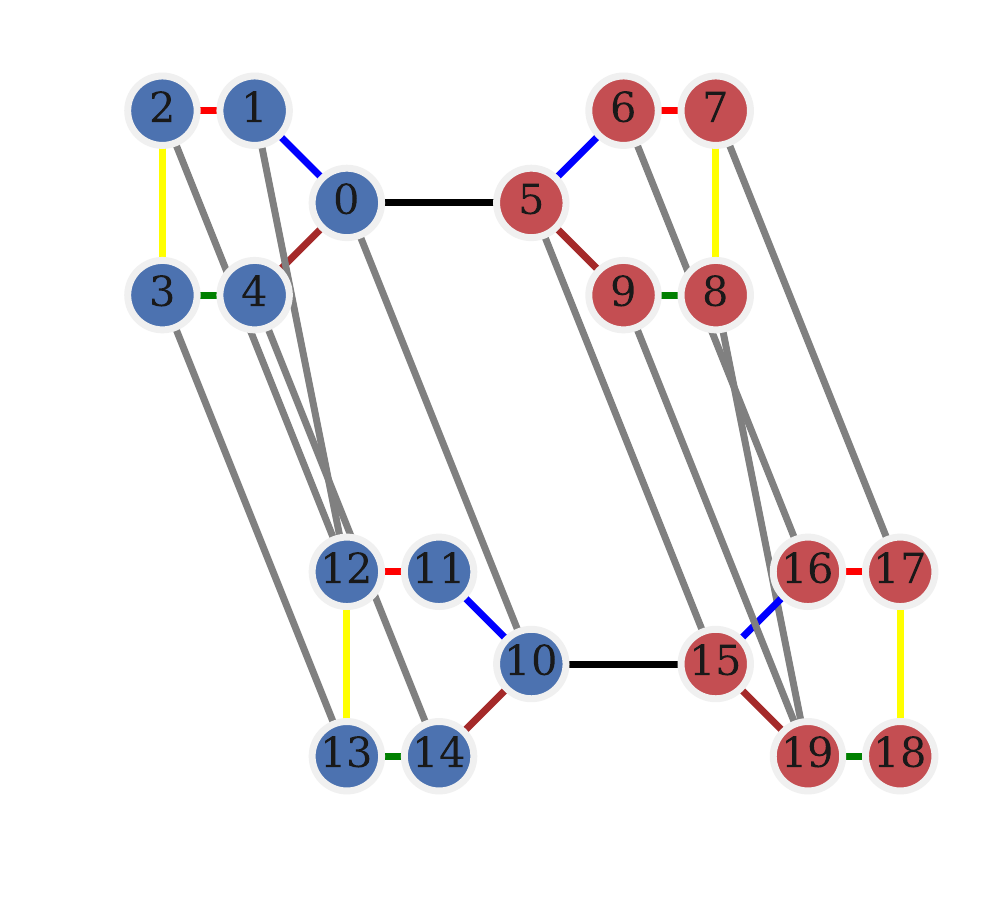}  \includegraphics[scale=0.25]{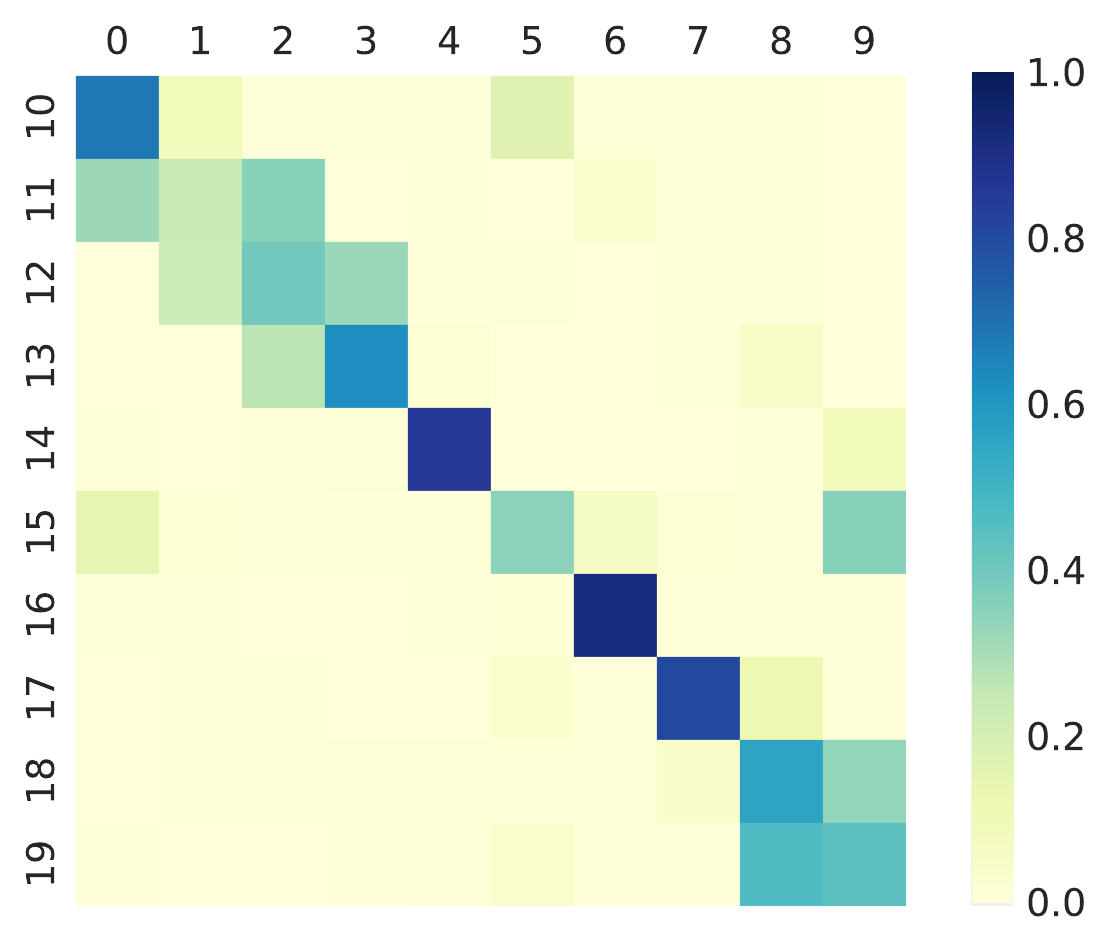}
        \caption{Labeled nodes and edges.}
        \label{fig:graph_edge_node_reg}        
    \end{subfigure}
    \caption{The effect of attributes preserving losses on the attention networks.
    Our method is given a pair of identical graphs, the upper graph represents the target and the other represents the source graph.
    Each graph consists of two rings of size $5$ connected with one edge (($0$, $5$) and ($10$, $15$) respectively).
    We visualize the strongest attention weights as cross-graph edges.
    On the right of each figure we visualize the rest of the attention weights as a heatmap.
    When the graph attends to itself without attribute preserving losses, there are several solutions that are equally good because of several symmetries available.
    Once we add the nodes attributes, we can see an immediate effect where the nodes from the same label class attend only to each other. This behavior further intensifies after also adding the edge attributes.
    }
    \label{fig:attention_viz}     
\end{figure*}

\subsection{Training}
\label{sec:training}
We implement our models using TensorFlow \cite{tensorflow},
calculate our gradients using backpropagation, and update our parameters using Adam \cite{kingma2014adam}.
We train each source graph on its adjacency matrix for a constant number of iterations.

\subsubsection{Target Graph Encoding}
Here, the augmented encoder has to predict the neighboring vertices for each vertex in the target graph with the help of the attention and reverse-attention layers.
To learn the augmented target graph encoder (which consists of the source graph encoder with the additional attention layers), we use the following procedure:
\begin{enumerate}
    \item First, freeze the parameters of the source graph encoder.
    \item Second, add two additional networks, one for attention and another for reverse attention mapping between the target graph nodes to the source graph nodes and vice versa.
    \item Third, add the regularizing losses to preserve the nodes or edges attributes if they are available.
    \item Fourth, train the augmented encoder on the input, which is the adjacency matrix of the target graph, and a node attribute and/or edge attribute matrix (if available).
\end{enumerate}
\noindent Finally, once the training of the attention layers is done, we use the augmented encoder to compute the divergence between the graph pair as discussed in \ref{sec:div}.

\subsection{Scalability}
\label{sec:scalability}
We start by defining the following quantities:
$N$ the number of source graphs in the dataset,
$M$ the number of target graphs in the dataset,
$V$ the average number of nodes,
$\tau{}$ the number of epochs to encode source graphs,
$\rho{}$ the number of epochs to encode target graphs,
$l$ the number of encoder hidden layers,
$m$ the number of attention hidden layers, and
$d$ the embedding and hidden layer size

Our method relies on pairwise similarity, therefore, we will have $M\times N$ computations that each involves scoring a target graph against one source graph.
Training a source graph encoder requires $\tau{}$ steps that each involves $2\times V\times d + l\times d^2$ computations.
In addition to running the source graph encoder, the target graph alignment learns the attention networks which represents $\rho{} \times (2\times d\times V + m \times d^2)$.
If we define $T = max(\rho{}, \tau{})$, $k = max(l, m)$, and $M=N$
then the total computation cost is 
$\Theta(N^2 \times T \times (V\times d + k \times d^2)$.
Because $V$ is likely much larger than $d^2$, we interpret the computational complexity as $O(TN^2V)$.

In Section \ref{sec:sampling}, we explore the effect of sampling to hasten \ours{}'s runtime on large datasets.  
We show that not all $M\times N$ comparisons are necessary to achieve high performance: empirically, it seems that less than $20\%$ of source graphs are required, significantly speeding our approach.

\section{Experiments}
In this section, we demonstrate our method through a series of qualitative and quantitative experiments.
First, we show how our attention based alignment works under different conditions.
Then, we show how our representations are capable of capturing the structure of the space of graphs by applying hierarchical clustering on the kernel space.
Finally, we show that the learned graph embeddings represent a sufficient feature set to predict the graph label on several challenging classification problems in real-world biological datasets.

\subsection{Cross-Graph Attention}

In this qualitative experiment, we seek to understand how two graphs are related to each other.
Comparing different patterns between different graphs is an important application in domains such as biology.

Figure \ref{fig:graph_no_reg} shows two identical unlabeled barbell graphs.
Each graph consists of two rings of size $5$ connected with the edge ($0$, $5$) and ($10$, $15$).
The upper graph represents the target graph while the lower one represents the source graph.
The edges connecting the source and target graphs represent the strongest attention weights for each node in the target graph.
The heatmap shows the full attention matrix for more thorough analysis.
Aligning these identical graphs is an easy task for the naked eye.
However, our method can find many possible symmetries to exploit while still achieving perfect predictions.
For example, nodes in the left ring can attend to the right ring of the source graph and vice versa.

Figure \ref{fig:graph_node_reg} shows the previous setup with labeled graph nodes.
This introduces a regularizing loss to preserve the node attributes.
The attention heat map shows significant weights for the upper left and lower rights quadrants.
The right ring does not attend to nodes in the left ring anymore, and vice versa.
Still we can see the method exploiting symmetries within the same ring.

Finally, by also adding edge labels, the alignment problem is constrained enough that the attention heatmap is concentrated along the diagonal (See Figure \ref{fig:graph_edge_node_reg}).
We can observe that the attention edges correspond in a one-to-one relationship between the target and source graphs.
This synthetic experiment shows the effect of attribute preserving losses on learning the alignment between graphs.

\subsection{Hierarchical Clustering}
To understand the global structure of the graph embedding space, we explore it qualitatively using hierarchical clustering.
First, we create a dataset which is a composition of 6 different families of graphs.
Three graph families are mutated graphs and three families are subset of a larger set of realistic graphs.
From each family we sample 5 graphs, creating a universe of 30 graphs.
Then, we embed the graphs using our method constructing a graph embedding space.
Finally, we cluster the embeddings according to their pairwise euclidean distances.

\subsubsection{Mutated Graphs}
For these datasets we start with a known graph and generate a sequence of mutations to produce a family of graphs.
In particular, we consider the following graphs.

\begin{itemize}
     \item C. Elegans \cite{watts1998collective}: represents the neural network of the C. Elegans worm.
     \item Karate Club \cite{ZacharyKarate}: social network of friendships between 34 members of a karate club.
     \item Word Network \cite{NewmanWordNetwork}: adjacency network of common adjectives and nouns in the novel David Copperfield by Charles Dickens.
 \end{itemize}

In order to generate a family $G_1 \cdots G_k$ for each original graph $G_0$, we employ the following mutation procedure.
At each of the $k$ time steps, there is a $p=0.5$ chance of performing an edge deletion or addition.
For additions, we select the two nodes to connect from any unlinked nodes according to the preferential attachment model characterized by $G_0$ \cite{chung2006complex}.
For deletions, we select an edge at random and remove it.
We run this procedure for 4 times with $k=50$ time steps, creating a family of 5 related graphs.
The initial seed for any of these mutations is denoted by the suffix ``-0".

\subsubsection{Realistic Graphs}
We randomly pick 5 graphs from three of the real-world families of graphs we consider (\textbf{D\&D}, \textbf{PTC}, and \textbf{NCI1}).
See Section \ref{sec:datasets} for more information about these graphs.

\begin{figure}
    \centering
    \includegraphics[width=\columnwidth]{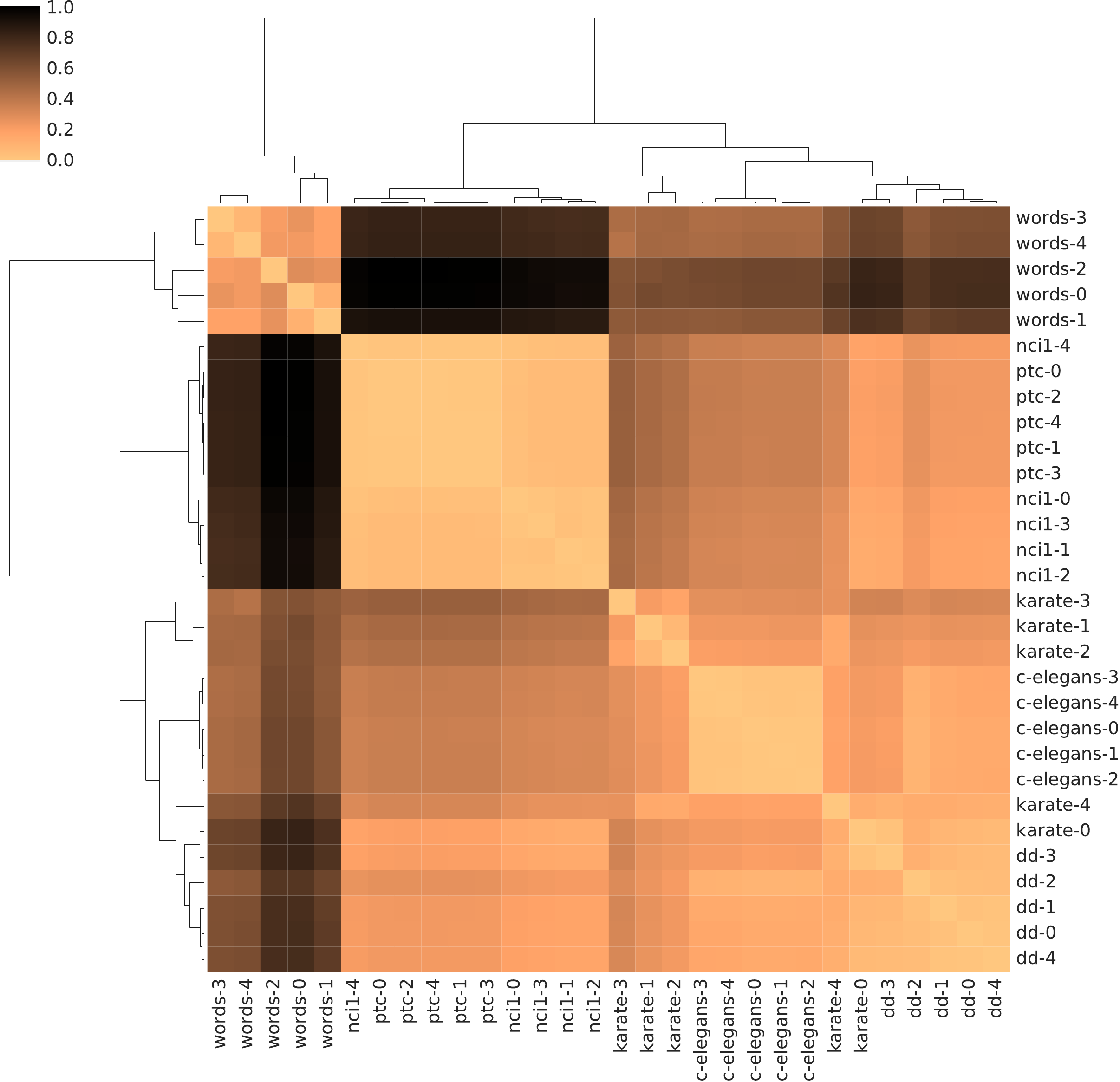}
    \vspace{3mm}
    \caption{A hierarchical clustering of the graph kernel space for several different graph families.  It shows 30 graphs that belong to 6 different families.
    The values of the matrix are the pairwise Euclidean distances between the graph embeddings.
    }
    \label{fig:clusters30}
\end{figure}

Figure \ref{fig:clusters30} shows the result of clustering the pairwise distances between our graph embeddings.
We are able to retrieve perfect clusters of \{\texttt{c-elegans}, \texttt{words}\} where there are clusters of size of 5 that consist only of graphs of the same type.
For \{\textbf{NCI1}, \textbf{D\&D\}}, we can cluster 4 graphs out of 5 before adding a graph which is out of the family.

\subsection{Graph Classification}

%\todo{Dustin: make splits available to public}

Our learned graph representations respect both attributes and structure.
They can be used for graph classification tasks where the graph structure, node attributes, and/or edge attributes convey meaning or function.
To demonstrate this, we use \ours{} representations of several chemo- and bio- informatics datasets as features for classification tasks.
We report our results against both unsupervised and supervised methods.

%For datasets with more than 2,000 graphs, we restrict the grid search to only C parameters.

%Input features are the DDGK graph representations: the final pair-wise scoring loss of all graphs in the dataset.
\subsubsection{Hyper-parameters Search}
To choose \ours{} hyperparameters (See Table \ref{table:ddgkparams}), we perform grid searches for each dataset.
We create splits of each dataset to avoid over-fitting, they are: \{\texttt{train}, \texttt{dev}, \texttt{test}\}.
We use the scikit-learn SVM \cite{pedregosa2011scikit} as our classifier, and we vary the kernel choices between \{\texttt{linear}, \texttt{rbf}, \texttt{poly}, \texttt{sigmoid}\} and the regularization coefficient $C$ between $10$ and $10^9$.
We choose the hyper-parameters of both \ours{} and the classifier that maximize the accuracy on the \texttt{dev} dataset.

\begin{table}[tb]
\begin{tabular}{l|l}
\textbf{Hyper-Parameter}& \textbf{Values}\\\hline
Node embedding & $2, 4, 8, 16, 32$ \\ 
Encoder layers & $1, 2, 3, 4$ \\
Learning rate & $10^{-4}$, $10^{-3}$, $10^{-2}$, $10^{-1}$, $1$ \\
Encoding epochs & $100, 300, 600$ \\
Scoring epochs & $100, 300, 600$ \\
Node preserving loss coefficient & $0, 0.25, 0.5, 1.0, 1.5, 2.0$ \\ 
Edge preserving loss coefficient & $0, 0.25, 0.5, 1.0, 1.5, 2.0$ \\ 
\end{tabular}
\vspace{3mm}
\caption{Values used during our grid search for  \ours{} graph representations learning hyper-parameters.}
\label{table:ddgkparams}
\end{table}

\begin{table}[tb]
\begin{tabular}{l|rrccrc}
&\thead{\bf \#\\\bf Graphs}
&\thead{\bf Average\\\bf Nodes}
&\thead{\bf Average\\\bf Edges}
&\thead{\bf \#\\\bf Labels}
&\thead{\bf \# Node\\\bf Labels}
&\thead{\bf \# Edge\\\bf Labels}
\\\hline
% ENZYMES & $600$ & $33$ & $62$  & $6$  & $3$ & $-$ \\ 
D\&D & $1178$ & $284$  & $716$  & $2$ & $89$ & $-$ \\
NCI1 & $4110$ & $30$ & $32$ & $2$ & $37$ & $-$ \\
% NCI109 & $4127$ & $30$ & $32$ & $2$ & $38$ & $-$ \\
PTC & $344$ & $14$ & $15$ & $2$ & $18$ & $4$ \\
MUTAG & $188$ & $18$ & $20$ & $2$ & $7$ & $4$ \\
\end{tabular}
\vspace{3mm}
\caption{Statistics of the chemo- and bio-informatics datasets.}
\label{table:datasetprop}
\end{table}

\begin{table*}[ht]
\begin{tabular}{ll|lll|llll}
\textbf{Method} & &\rotatebox[origin=l]{90}{\tiny \bf Unsupervised}&\rotatebox[origin=l]{90}{\tiny \bf No Weisfeiler-Lehman}&\rotatebox[origin=l]{90}{\tiny \bf No Feature Engineering} &\textbf{D\&D}&\textbf{NCI1}&\textbf{PTC}&\textbf{MUTAG}\\\hline
% These results are GNNs that try to encode specific problem domain information

% These are the results for PSCN where k=10
% dataset MUTAG PCT NCI1 PROTEIN D & D
% PSCN k=10 88.95 ± 4.37 (3s) 62.29 ± 5.68 (6s) 76.34 ± 1.68 (76s) 75.00 ± 2.51 (30s) 76.27 ± 2.64 (154s)
PSCN & \cite{patchysan} &&&  & $76.27\pm2.64$ & $76.34\pm1.68$ & $62.29\pm5.68$  & $88.95\pm 4.37$  \\
DGCNN & \cite{zhang2018end} &&&  & $-$ & $74.44\pm0.40$ & $58.59\pm2.40$  & $85.83\pm1.60$  \\

% % These are semi-supervised methods
% graphsage & & &&&  $54.25$ & $75.42$ & $-$ & $-$ & $-$  & $-$  \\
% \hline

% % I'm not sure what these are yet

% diffpool & & &&&  $\textbf{64.23}$  & $81.15$ & $-$ & $-$ & $-$  & $-$  \\
% \hline

\hline

% These results are the simplest baselines - just basic kernels
%SP Kernel & (PSCN team) &&& & $-$ & $-$ & $73.00\pm0.51$ & $-$ & $58.53\pm2.55$ & $85.79\pm2.51$\\
SP Kernel & \cite{borgwardt2005shortest} &\cmark&&  & $79.00\pm0.60$  & $74.50\pm0.30$ & $58.90\pm2.20$  & $83.00\pm1.40$ \\  
% RW Kernel & (PSCN team) &&&  & $-$ & $-$ & $-$ & $-$ & $57.26\pm1.30$  & $83.68\pm1.66$ \\
% WL Kernel & (PSCN team) & \cmark &&  & $-$ & $77.95\pm0.70$  & $80.22\pm0.51$ & $-$ & $56.97\pm2.01$  & $80.72\pm3.00$  \\
WL Kernel & \cite{kriege2016valid} & \cmark &&  & $79.00\pm0.40$  & $85.80\pm0.20$ & $61.30\pm 1.40$  & $86.00\pm1.70$  \\
% Optimal assignment WL algorithm
WL-OA & \cite{kriege2016valid} & \cmark && & $79.20\pm0.40$ & $\textbf{86.10}\pm0.20$ & $63.60\pm1.50$  & $86.00\pm1.70$  \\

DGK & \cite{yanardag2015deep} & \cmark &&  & $-$ & $80.30\pm0.40$ & $60.10\pm2.50$  & $87.40\pm2.70$  \\

graph2vec & \cite{graph2vec} & \cmark &&  & $-$ & $73.22\pm1.90$ & $60.17\pm6.90$  & $83.15\pm9.20$  \\

S2S-N2N-PP & \cite{taheri2018RNN} & \cmark && & $-$ & $83.72\pm0.40$ & $\textbf{64.54}\pm1.10$ & $89.86\pm1.10$  \\
\hline
% Supervised Classifier on Unsupervised Representation
node2vec & \cite{node2vec} & \cmark  & \cmark &  & $-$  & $61.91\pm0.30$ & $55.60\pm1.40$  & $82.01\pm1.00$  \\

\hline
\ours{} & (this paper)  & \cmark & \cmark & \cmark&  $\textbf{83.14}\pm2.72$ & $68.10\pm2.30$  & $63.14\pm6.57$ & $\textbf{91.58}\pm6.74$ \\
\end{tabular}
\vspace{3mm}
\caption{Average accuracy in ten-fold cross validation on our graph classification task. 
Methods are grouped by their level of supervision during the similarity metric learning, whether they use algorithm insights the Weisfeiler-Lehman algorithm, and  whether they use feature engineering (e.g. graph motifs, random walks, etc.).
Baseline results taken from \cite{patchysan,kriege2016valid,taheri2018RNN} (missing results are missing from these works).
We note that \ours{} performs surprisingly competitively for an unsupervised method with no hard-coded insights.
}
\label{table:results}
\end{table*}

\subsubsection{Datasets}

\label{sec:datasets}
Four benchmark graph classification datasets from chemo- and bio-informatics domains are used. The datasets include \textbf{D\&D}, \textbf{NCI1}, \textbf{PTC} and \textbf{MUTAG}.
All datasets include node labels.
The \textbf{PTC} and \textbf{MUTAG} datasets also include edge labels.
Table \ref{table:datasetprop} shows network statistics for each dataset.
The datasets:
\begin{itemize}
% The \textbf{ENZYMES} dataset contains 600 enzymes from the \textbf{BRENDA} \citep{Schomburg2004} enzyme database. The enzymes are labeled with one of the six top-level Enzyme Commission numbers. 
\item \textbf{D\&D} \citep{dobson2003distinguishing}:   contains 1178 proteins labeled as enzymes or non-enzymes.
\item \textbf{NCI1} \citep{45b3e5c6d2ee4938b77995a88ee0b928}: contains 4110 chemical compounds labeled as active or inactive against non-small cell lung cancer.
\item \textbf{PTC} \citep{toivonen2003statistical}: contains 344 chemical compounds labeled according to their carcinogenicity in male rats.
\item \textbf{MUTAG} \citep{doi:10.1021/jm00106a046}:  contains 188 mutagenic aromatic and heteroaromatic compounds labeled according to their mutagenic effect on a specific gram negative bacterium.
\end{itemize}

\subsubsection{Results}

The results of these experiments are presented in Table \ref{table:results}.
We see that \ours{} is quite competitive, with higher average performance on both the \textbf{D\&D} and \textbf{MUTAG} datasets than any of the baselines.  
This is especially surprising given that the supervised methods have additional information available to them.
We note that \ours{} achieves its strong results without engineered features, or access to information from Weisfeiler-Lehman kernels.
For \textbf{PTC}, we also see that \ours{} attains competitive performance against all other methods, only being outperformed by 2 of the 9 baselines.
Finally, on \textbf{NCI1}, we see that \ours{} performs better than the method using the most similar kind of information (\texttt{node2vec}), but find that baselines using the WL kernel perform best on this dataset (indeed, the WL kernel itself takes the top two spots).
We find this dependence quite interesting, and will seek to characterize it better in future work.

\subsection{Dimension Sampling}
\label{sec:sampling}
So far, we have been setting the source graphs set to be equal to the target graphs set.
This pairwise computation is quite expensive for large datasets.
To reduce the computational complexity of our method, we study the effect of sub-sampling the dimensions of our graph embedding space on the quality of graph classification.

\begin{figure}
    \centering
    \includegraphics[width=\columnwidth]{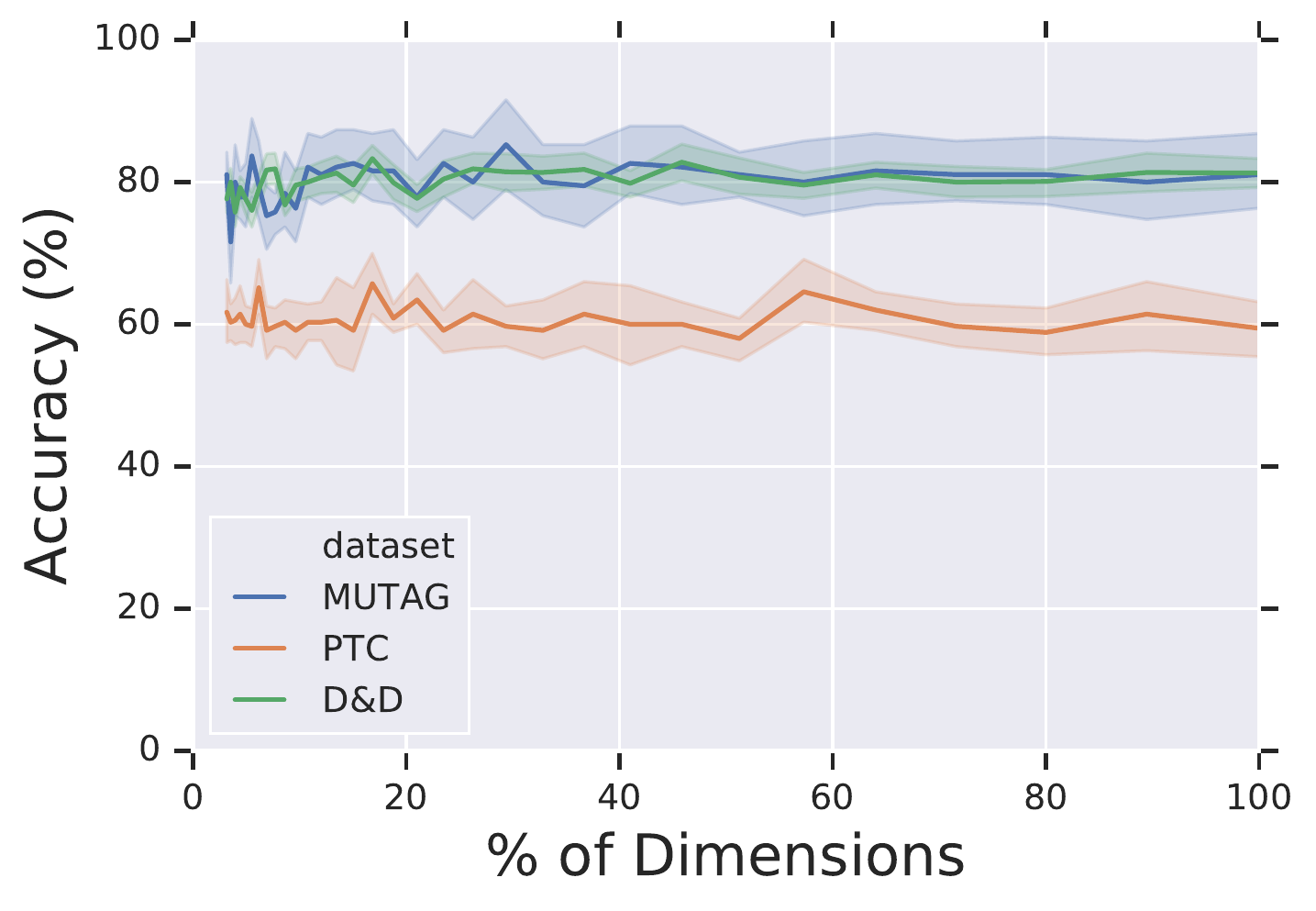}
    \caption{Effect of sub-sampling source graphs on graph classification tasks.  Here we vary the number of source graphs available to each method, and observe that very few dimensions are needed to achieve our final classification performance (less than $<20\%$ of the dimensions for the datasets considered).}
    \label{fig:sampling}
\end{figure}

To do that, we  construct a source graph set that is a subset of the original graph set.
We learn divergence scores for all target graphs against this subset.
We use the reduced embeddings as features to predict graph categories.
Figure \ref{fig:sampling} shows that we are able to achieve stable and competitive results with less than 20\% of the graphs being used as source graphs.

\section{Related Work}

The main differences between our proposed method and
previous work can be summarized as follows:
\begin{enumerate}
\item We are an unsupervised method, taking only a graph as input.
\item We use no domain-specific information about what primitives are important in a graph, using only the edges.
\item We use no algorithmic insights from the literature in graph isomorphism (e.g.\ the Weisfeiler-Lehman kernel).
\item We assume nothing about the mapping of node ids between graphs, instead learning the alignment.
\end{enumerate}

\noindent While many approaches exist that contain at least one of these differentiators, we are, to the best of our knowledge, the only proposed method that meets all four of these conditions. 
In this section we will briefly cover related work in graph similarity and other applications of neural networks to graph representation.

\subsection{Unsupervised Graph Similarity}

We divide our brief survey of the literature into three kinds of unsupervised methods for graph similarity.
The first seeks to explicitly define a kernel over graph features, or use the intuition from such a kernel as part of the representation learning process.
The second focuses on the representation of individual elements of the graph, learning primitives that maximize some kind of reconstruction of the graph.  
The third group of work constructs a similarity function between graphs by an explicit vector of statistical features constructed by the graph.

% unsupervised graph kernels

% \noindent \textbf{Unsupervised Methods}:
\noindent \textbf{Traditional Graph Kernels}:
There has been considerable work done on unsupervised methods for graph kernel learning.
Initial efforts in the area focused on theoretical views of the problem, defining graph similarity via the Graph Edit Distance \cite{gao2010survey} or the size of the Maximum Common Subgraph \cite{bunke2002comparison} between graphs.  Unfortunately these problems are both NP-Complete in the general case, require a known correspondence between the nodes of the two graphs of interest.

Many approaches are built around the graph similarity measure computed by the Weisfeiler-Lehman (WL) subtree graph kernel \citep{shervashidze2011weisfeiler,kriege2016valid}.  
At its core, the WL algorithm collapses the labels of a node's neighbors into a ordered sequence, and then hashes that sequence into a new label for the node.
This process repeats iteratively to average information over the neighborhood together.
Other functions that use different types of predefined features for graph similarity, such as shortest-paths kernels \cite{borgwardt2005shortest}, and random walk kernels \cite{kashima2003marginalized} have also been proposed, but their naive implementations suffer from high asymptotic complexity ($O(n^4)$ and $O(n^6)$, respectively).
Faster implementations of these kernels have been proposed \cite{borgwardt2007fast,kang2012fast}.
Some unsupervised methods also focus on extending the algorithm intuition of these classic approaches to the problem.
For instance \cite{taheri2018RNN} learns a representation for each position in a WL ordering jointly while learning a graph representation.

Unlike all of these approaches, our method deliberately avoids algorithmic insights.  
Our proposed isomorphism attention mechanism allows capturing higher-order structure between graphs (beyond immediate neighborhoods).

\noindent \textbf{Node embedding methods}: Since DeepWalk \cite{deepwalk} proposed embedding the nodes via a sequence of random walks, the problem of node representation learning has received considerable attention \cite{perozzi2017don,node2vec,chen2017harp,tsitsulin2017verse,bojchevski2017deep,abu2018watch,aepasto2019}.  In general, all of these methods utilize insights about similarity functions which are important to the graph.
While these methods seek the best way to represent nodes, the representations are learned independently between graphs, which makes them generally unsuitable for graph similarity computations.
For more information on this area, we recommend recent surveys in the area \cite{chen2018tutorial,cui2018survey}.
Unlike these methods, our goal is to learn representations of graphs, not of nodes.

\noindent \textbf{Graph statistics}:
Finally, another family of unsupervised graph similarity measures define a hand-engineered feature vector to compute graph similarity.  
The NetSmilie method \cite{berlingerio2012netsimile} operates by constructing a fixed size feature value of graph statistics and uses this as a similarity embedding over graphs.  
Similarly, DetlaCon \cite {koutra2013deltacon} defines the similarity over two graphs with known node-to-node mapping via the similarity in their propagation of belief, and \cite{papadimitriou2010web} proposes a number of similarity measures over directed web graphs.

Unlike these methods, \ours{} does not explicitly engineer its features for the problem.  Instead, the similarity is learned function directly from the edges present in the adjacency matrix, with no assumptions about which features are important for the application task.

% supervised graph embedding approaches
\subsection{Supervised Graph Similarity}
%\noindent \textbf{Semi-supervised Methods}:
The first class of supervised methods uses some supervision to inform a similarity function constructed over different hand-engineered graph features.

A number of supervised approaches also utilize intuitions from the Weisfeiler-Lehman graph kernel.
\texttt{Patchy-SAN} \citep{niepert2016learning} proposes an approach for convolutional operations on graph structured data.  The core of their method uses the ordering from the WL kernel to order the nodes of a rooted subgraph into a sequence, and then apply standard 1-dimensional convolutional filters.
This approach is further generalized by \citep{zhang2018end}, who use the WL ordering to sort a graph sample in a pooling layer.
Another branch of work has focused on extending the Graph Convolutional Networks (GCNs) proposed by \cite{kipf-gcn} to perform supervised classification of graphs.
Proposed extensions include a pooling architecture that learns a soft clustering of the graph \citep{ying2018hierarchical}, or a two-tower model which frames graph similarity as link prediction between GCN representations \cite{bai2018graph}.
Interestingly, it has been shown that many of these methods are not necessarily more expressive than the original Weisfeiler-Lehman subtree kernel itself \cite{morris2018weisfeiler}.

Unlike all of these approaches, our method learns representations of graphs without supervision --- we use no labels about the class label of a graph, and no external information about which pairs of graphs are related.
Our proposed isomorphism attention mechanism allows capturing higher-order structure between graphs (beyond immediate neighborhoods).

% \subsection{Representation Learning}
% % node embedding approaches
% The embedding of the nodes in a graph has received considerable attention  \cite{deepwalk}.

\section{Extensions \& Future Work}
\label{sec:extensions}
Here we briefly discuss a number of areas of future investigation for our method.

\subsection{Graph Encoders}
Given the choice of input and reconstructed output, several additional graph encoders are possible, in addition to the Nodes-To-Edges encoder which we used in this work.  To mention a few options:

\paragraph{Edge-To-Nodes Encoder} -
This encoder is trained to predict the source and destination vertices given a specific edge.
Similar to the \emph{Node-To-Edges} encoder, this could be expressed as a multilabel classification task with the following objective function,
\begin{equation}
J(\theta) = \sum_{e_{ij} \in E} \log \Pr(v_i \mid e_{ij}, \theta) + \log \Pr(v_j \mid e_{ij}, \theta)
\end{equation}
Note that the number of edges in a graph could grow quadratically, therefore, iterating over the edges is more expensive than the nodes.

\paragraph{Neighborhood Encoder} -
In this case, the encoder is trained to predict a set of vertices or edges that are beyond the immediate neighbors.
Random walks could serve as a mechanism to calculate a neighborhood around a specific node or edge.
Given a partial random walk, the encoder has to predict the vertices that could have been visited within a specific number of hops.

\begin{equation}
   J(\theta) = \sum_{\substack{(v_1, v_2, \cdots, v_{i}) \\ \thicksim RandomWalk(G, E, V)}} \log \Pr\big(v_{j}\mid( v_1, v_2, \cdots, v_{i}, \theta)\big)
\end{equation}

\subsection{Attention Mechanism}
We proposed a simple attention mechanism which uses node-to-node alignment.
As we discussed in Section \ref{sec:scalability}, we could replace the linear layer with a deep neural network to reduce the size of the model if scability is an issue.
While node-to-node alignment enhances the interpretability of our models, subgraph alignment could lead to better and easier understanding of how two graphs are similar.
Hierarchical attention models \cite{yang2016hierarchical} could lead to higher levels of abstractions which could learn community structure and which communities are similar across a pair of graphs.
Hierarchy has already been used within the context of learning better node embeddings, for example \citep{ying2018hierarchical} showed that a better understanding of the graph substructure can lead to better graph classification.
Therefore, we believe extending the work beyond node-to-node alignment will significantly improve our results.

\subsection{Regularization}
We proposed attribute based losses to regularize our isomorphism attention mechanism.
The graph encoder capacity was adjusted according to the source graph size.
However, the source graph encoder could still suffer from overfitting which would reduce its utility in recognizing similar target graphs.
Therefore, further research is necessary to understand the relation between the encoder training characteristics and the quality of the generated divergence scores

\subsection{Feature Engineering}
In this work we have focused on developing an approach for representing graphs that operated without any feature engineering or algorithmic insights.
While this willful ignorance has allowed us to design a new paradigm for graph similarity, we suspect that there are many fruitful combinations of this idea with other approaches for graph classification.
For example, the graph embeddings we learn could be used as additional features for approaches based on learning supervised classifiers over graphs.

\section{Conclusion}
In this work, we have shown that neural networks can learn powerful representations of graphs without explicit feature engineering.
Our proposed method, Deep Divergence Graph Kernels, learns an encoder for each graph to capture its structure, and uses a novel \textit{isomorphism preserving attention mechanism} to align node representations across graphs without the use of supervision.
We show that representing graphs by their divergence from different source graphs provides a powerful embedding space over families of graphs.
Our proposed model is both flexible and amenable to extensions.
We illustrate this by proposing extensions to handle many commonly occurring varieties of graphs, including graphs with attributed nodes, and graphs with attributed edges.

% The quality of the augmented encoder predictions defines a divergence score for each pair of graphs.
% Finally, We construct an embedding space for all graphs using the pair-wise divergence scores. 

Our experimental analysis shows that despite being trained with only the graph's edges (and no feature engineering) the learned representations encode a variety of local and global information.
When the representations produced by \ours{} are used as features for  graph classification methods, we find them to be competitive with challenging baselines which use at least one of graph labels, engineered features, or the Weisfeiler-Lehman framework.
In addition to being powerful, \ours{} models are incredibly informative.
The learned isomorphism attention weights allow a level of insight into the alignment between a pair of graphs, which is not possible with other deep learning methods developed for graph similarity.

Unsupervised representation learning for graphs is an important problem, and we believe that the method of Deep Divergence Graph Kernels we have introduced here is an exciting step forward in this area.
As future work, we will investigate 1) enhanced method for choosing informative source from the space of all graphs, 2) improving the architecture of our encoders and attention models, 3.) making it easier to reproduce research results in the area of graph similarity, and 4) making graph similarity models even easier to understand.

%\begin{acks}
%\end{acks}

\newpage
\bibliographystyle{ACM-Reference-Format}
\bibliography{references}

\end{document}